# 基于有限差分残差物理约束的波动方程无监督学习方法

冯鑫 [1,2]，姜 屹 [3]，秦嘉贤 [3]，张来平 [4]，邓小刚 [1,3]

（1. 四川大学 计算机学院，成都 610065；2.四川大学 天府工程数值模拟与软件创新中心，成都 610207；3.军事科学院 系统工程研究院，北京 100082；4.军事科学院 国防科技创新研究院，北京 100071）

**摘 要：** 波动方程是一种重要的物理偏微分方程,近年来深度学习有望加速或替代传统数值方法对其求解.然而现有深度学习方法存在着数据集获取成本高、训练效率低、边界条件泛化能力不足的问题,为此本文提出一种基于有限差分残差约束的波动方程无监督学习方法,我们基于结构网格和有限差分方法构建一种新颖的有限差分残差约束,以及一种无监督训练策略,使得卷积神经网络能够在无数据集条件下训练,并预测波的正演过程.实验结果表明,有限差分残差约束相较于 PINNs 类的物理信息约束具有更容易拟合、计算成本更低、源项泛化能力更强的优点,这使得我们的方法有着更高的训练效率和应用潜力.

**关键词：** 卷积神经网络；有限差分方法；波动方程；无监督学习

**中图分类号：** TP391　　　　　　　　　　**文献标识码：** A

## Unsupervised Learning Method for the Wave Equation Based on Finite Difference Residual Constraints Loss

FENG Xin[1,2],JIANG Yi[3],QIN Jia-Xian[3],ZHANG Lai-Ping[4],DENG Xiao-Gang[3]

（1. College of Computer Science, Sichuan University, Chengdu 610065, China; 2.Tianfu Engineering-oriented Numerical Simulation & Software Innovation Center, Sichuan University, Chengdu 610000, China; 3. Institute of Systems Engineering, Academy of Military Sciences(AMS), PLA , Beijing 100082, China; 4. Institute of Defense Science and Technology Innovation, Academy of Military Sciences, Beijing 100071, China）

**Abstract:** The wave equation is an important physical partial differential equation, and in recent years, deep learning has shown promise in accelerating or replacing traditional numerical methods for solving it. However, existing deep learning methods suffer from high data acquisition costs, low training efficiency, and insufficient generalization capability for boundary conditions. To address these issues, this paper proposes an unsupervised learning method for the wave equation based on finite difference residual constraints. We construct a novel finite difference residual constraint based on structured grids and finite difference methods, as well as an unsupervised training strategy, enabling convolutional neural networks to train without data and predict the forward propagation process of waves. Experimental results show that finite difference residual constraints have advantages over physics-informed neural networks (PINNs) type physical information constraints, such as easier fitting, lower computational costs, and stronger source term generalization capability, making our method more efficient in training and potent in application.

## 1 引 言

波是一种重要的信息传播方式,它在自然界和技术应用中有着重要作用,波动方程能够描述各种类型的波,包括机械波、电磁波、声波等,因此对于许多物理问题而言求解波动方程是一个关键任务.在声学、地球物理学、电磁学、流体动力学等领域有许多波动方程的研究工作,例如对地震源进行定位[1],估计地下结构[2],电磁场模拟[3],气动噪声模拟[4],流体噪声模拟[5]等.

过去几十年,人们发展了数值方法求解波动方程,例如有限差分方法(Finite-difference methods, FDM)、有限元法(Finite element method, FEM)、谱元法(Spectral element method, SEM)等.这些方法基于网格将波动方程离散化,并使用迭代的时间步进格式求解解它.基于过去几十年的发展,数值方法能够解许多复杂的问题,例如在声学中 jiang 等人针对一个襟翼外形的大涡模拟辨识其中不同的声源[6],在地球物理学中,yin 等人采用优化差分算子解决高精度频率域弹性波方程的数值频散问题[7],然而数值方法主要

缺点是计算成本高昂,对于一些复杂的三维问题,往往需要超级计算机来运行计算程序[8],[9],这需要耗费成千上万的计算核时,因此研究如何加速数值方法有着重要意义.

近年来深度学习对于物理现象模拟有极好的应用前景.例如 Sanchez-Gonzalez 等人提出了一种基于图神经网络的机器学习框架用以模拟流体、刚性固体和可变形材料相互作用[10],并结合了网格离散化提高了框架的训练效率和扩展能力.基于数据驱动的方法在波传播预测有许多工作,Azulay 等人提出一种多尺度的数据驱动方法用于迭代求解高波数下的离散异质亥姆霍兹方程[12].Gantala 等人提出了一种基于卷积长短期记忆(ConvLSTM)架构的 DPAI 模型模拟超声波在二维空间中的传播[13],Moseley 等人工作表明深度神经网络可以模拟波在复杂介质中的传播,并且速度相较于有限差分方法快 20-500 倍.然而上述的数据驱动方法普遍存在两个问题,一是方法需要依赖传统数值方法以构建训练数据集;二是神经网络在训练数据之外的计算场景表现不佳 [14].

物理信息机器学习[15]在近年来逐渐兴起,不同于纯数据驱动训练神经网络,这种方法将物理信息以约束形式混合到训练过程中,试图结合二者的优点.其中一个重要的工作是物理信息神经网络(physics-informed neural networks, PINNs),由 Raissi 等人[16],[17]提出.PINNs 是一种通用的偏微分方程求解方法,它将神经网络视作方程的解析解,基于自动微分技术和物理方程构建物理信息约束,这种约束描述了神经网络络对于方程解析解的拟合程度.物理信息约束允许神经网络在少量和无数据的条件下训练.Raissi 等人[17]表明 PINNs 可以预测非采样点的结果,这为神经网络的泛化能力提供了解决思路.此后有许多基于 PINNs 求解波动方程的工作被提出,Wang 等人使用 PINNs 对大地电磁场进行正向建模[18],Smith 等人基于 PINNs 求解 Eikonal 方程[19],Moseley 等人验证了均匀、分层和真实地球模型中 PINNs 求解波动方程的有效性[20].虽然相较于数据驱动方法,PINNs 有更好的泛化能力,但是 Song 等人发现当波动方程的边界条件发生变化时,模型需要重新训练,这限制了模型的应用场景,同时基于自动微分技术构建的物理信息约束带来了高昂的训练成本[21].

研究人员发现传统数值方法可以用于构造物理信息约束,这种数值物理约束可以将卷积神经网络更强的推理泛化能力和 PINNs 的训练策略结合起来,这可能是一种数值方法和物理信息学习更好的结合思路.其中在求解波动方程相关工作中,Wandel 等人[22]基于埃尔米特插值方法将卷积神经网络和物理信息神经网络的优点结合起来提出了 Spline-PINN 方法.Spline-PINN 的预测目标是样条系数,以构建方程解析解,这使得卷积神经网络络能够预测不同边界条件的波动方程的连续解,同时该方法基于物理信息约束实现了无监督训练.然而不直接预测方程的解使得

Spline-PINN 对于物理信息约束的收敛程度表现不佳,并且埃尔米特插值带来较大的性能开销导致了训练效率的下降.

为了改进现有物理信息方法在边界条件泛化、模型收敛精度,以及训练效率的不足,我们提出了一种基于有限差分残差约束的波动方程无监督学习方法.本文的贡献如下:(1)针对波动方程问题,提出一种有限差分残差约束,将传统有限差分方法和物理信息学习结合起来,解决了现有方法边界条件泛化能力不足,物理信息约束收敛程度表现不佳的问题.(2)提出一种基于训练池的无监督训练策略,使神经网络模型能够在无数据的条件下训练,提高了模型的训练效率.

## 2 基础知识

### 2.1 波动方程

波动方程描述了波的传播过程,在时域内不可压介质的波动方程如公式(1)所示:

$$
\begin{cases}
\dfrac{\partial u(\mathbf{X},t)}{\partial t} = -\dfrac{\partial p(\mathbf{X},t)}{\rho_0 \partial x} \\
\dfrac{\partial v(\mathbf{X},t)}{\partial t} = -\dfrac{\partial p(\mathbf{X},t)}{\rho_0 \partial y} \\
\dfrac{\partial p(\mathbf{X},t)}{\partial t} = -\rho_0 c^2 \left( \dfrac{\partial u(\mathbf{X},t)}{\partial x} + \dfrac{\partial v(\mathbf{X},t)}{\partial y} \right)
\end{cases}
\tag{1}
$$

式中 $t$ 是时间,$\rho_0$ 是传播介质密度,$c$ 是波速.$p(\mathbf{X},t)$ 是压力场,$u(\mathbf{X},t)$、$v(\mathbf{X},t)$ 是 $x$ 方向和 $y$ 方向的质点振速度,其中 $\mathbf{X} = \{x,y,z\}$,它表征了笛卡尔坐标系中的空间坐标.在现实中波是无限传播的,这与计算机有限的内存空间相矛盾,因此需要在计算域中设置吸收波边界条件[23],[24]或者完美匹配层[25],[26].在一个二维有界计算域中使用完美匹配层,二维波动方程可以表示为公式(2)所示:

$$
\begin{cases}
\dfrac{\partial u(\mathbf{X},t)}{\partial t} + \sigma * u = -\dfrac{\partial p(\mathbf{X},t)}{\rho_0 \partial x} \\
\dfrac{\partial v(\mathbf{X},t)}{\partial t} + \sigma * v = -\dfrac{\partial p(\mathbf{X},t)}{\rho_0 \partial y} \\
\dfrac{\partial p(\mathbf{X},t)}{\partial t} + \sigma * p = -\rho_0 c^2 \left( \dfrac{\partial u(\mathbf{X},t)}{\partial x} + \dfrac{\partial v(\mathbf{X},t)}{\partial y} \right)
\end{cases}
\tag{2}
$$

其中 $\sigma$ 是一个可微函数用于在完美匹配层中吸收波能.在完美匹配层区域外 $\sigma$ 等于零,在完美匹配层区域内,$\sigma$ 可以定义为如公式(3)[26]所示.式中 $\delta$ 是完美匹配层的层厚,$v$ 是波的传播速度,R 通常取 0.001,$l_{x,y}$ 表示位于 PML 内部的点 $(x,y)$ 到非 PML 区域的水平(x)和垂直(y)边界的边界的物理距离.

$$
\sigma = log \left( \dfrac{1}{R} \right) \dfrac{3v}{2\delta} \left( \dfrac{l_{x,y}}{\delta} \right)^2
\tag{3}
$$

### 2.2 交错网格

在使用有限差分方法求解波动方程时,通常采用交错网格(Staggered grid)对物理场的空间分布离散,这可以提高有限差分方法的稳定性和准确性.如图 1 所示,在二维空间

中，计算域被笛卡尔网格均匀离散为若干个计算单元，其中波动方程的物理量$(p,u,v)$交错分布在计算单元中。黄色计算单元和白色计算单元分别代表完美匹配层区域(PML Region)和模拟波传播的有界内部空间(Interior Region)，由完美匹配层函数$\sigma$的值划分。基于上述交错布局，物理量被均匀分布在计算域中，因此我们可以使用有限差分方法计算波动方程的空间偏导，同时构造适用于卷积神经网络的模型特征。

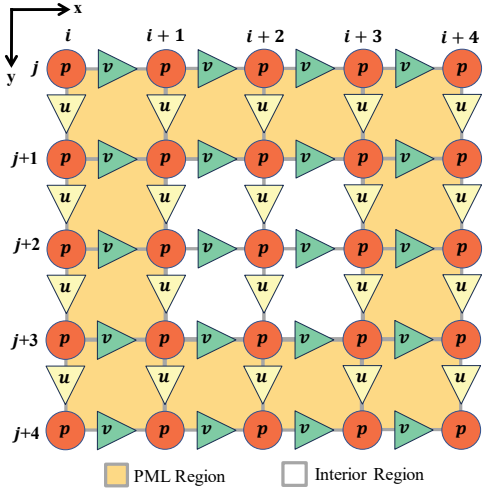

图 1 二维空间中的交错网格离散物理场

Fig. 1 Staggered mesh in 2D space

## 3 本文方法及模型

### 3.1 方法框架

如图 2 所示，我们的方法主要分为三部分：神经网络模型(Model)、物理信息(Physics Info)、训练池(Training Pool)。

在神经网络模型(Model)模块中，因为卷积神经网络(CNN)与有限差分方法(FDM)均具备平移不变形和局部相关性的内在归纳偏置，因此卷积神经网络能够利用笛卡尔网格提供的空间信息，以学习物理空间的时空变化。我们发现只需要简易结构(如图 4)的卷积神经网络即可取得较好的效果，在下文中简称为神经网络(NN)。

在物理信息(Physics Info)模块中，为了提高神经网络对于边界条件泛化能力不足，模型收敛程度不高的问题，我们基于有限差分卷积核和波动方程(Wave Equation)构造了一种新颖的有限差分残差约束(FDRC Loss)，通过将约束优化至 0 为神经网络模型提供监督信号。我们将在 3.2 节将介绍有限差分卷积核，并在 3.3 节讨论FDRC Loss 如何构造以及其相较于物理信息约束的优劣。

在训练池(Training Pool)模块中，为了解决数据集获取困难的问题，我们将模型训练过程中产生的迭代数据$(u^t, v^t, p^t)$更新到数据集(Wave Dataset)中，并从数据集中随机采样用于下一轮的训练，这使得训练过程中没有引入真实的标签解(Ground Truth)，实现了无监督训练。管理层(Manager)则采用了缓存和重置策略提高模型的训练效率。详细的训练策略将在 3.4 节介绍。

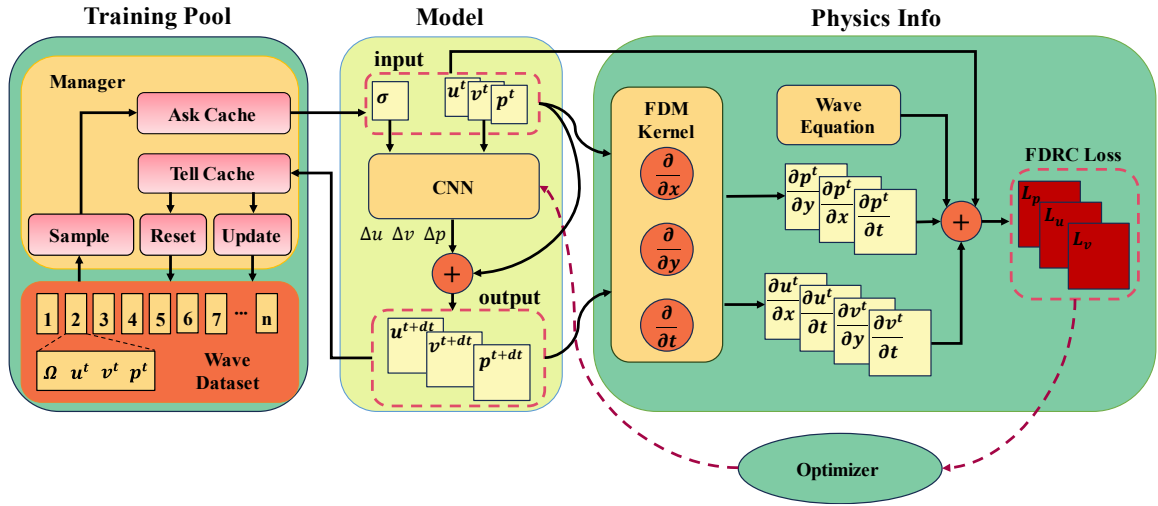

图 2 无监督训练框架

Fig. 2 Unsupervised Training Framework

### 3.2 有限差分卷积核

有限差分方法(Finite Difference Method, FDM)是一种数值分析和数值模拟技术，它将偏微分方程中的导数用差分逼近替代，从而将连续的问题转化为离散的问题，以近似原始方程的解。

我们使用了一种有限差分卷积核(FDM Kernel)去描述有限差分方法计算偏导项的离散项系数，通过该卷积核我们成功建立了神经网络与有限差分算子的梯度关系，这使神经

网络能学习有限差分方法的内在映射关系.例如图 3 所示,对于函数 $y = x^2 + 1$ 在 $x \in [0, 0.4]$ 的计算域上取 $dx = 0.1$ 进行离散,当使用前向差分(Forward Difference)计算一阶偏导项 $dy$ 时,其差分格式可以描述为 $dy = \frac{-y(x) + y(x + dx)}{dx}$,因此对于离散点 $\{x - dx, x, x + dx\}$ 其离散计算系数可以描述为 $\{0, -\frac{1}{dx}, \frac{1}{dx}\}$,也即有限差分卷积核的权重.基于这种思路,可以构建其他差分格式的有限差分卷积核.

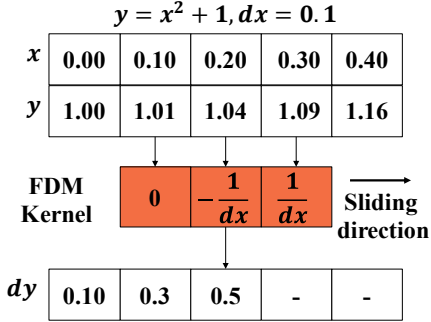

图 3 有限差分卷积核

Fig. 3 FDM Kernel

### 3.3 有限差分残差约束

本节将以二维空间中波动方程作为具体例子讨论有限差分残差差物理约束(Finite Difference Residual Constraints Loss, FDRC Loss)如何构建以及该约束相较于物理信息约束的优缺点.

基于波动方程构建监督信号以训练神经网络,在文献中有许多工作,其中主要是基于物理信息神经网络(physics-informed neural networks, PINNs).这些方法的约束是隐式的,它们通过隐式场描述的导数计算损失,使神经网络逼近微分方程在空间和时间上的高阶连续性.因为神经网络的拟合目标是对特定边界条件的方程解析解,当方程边界条件发生变化时就需要对神经网络重新训练,这些限制了方法的实际应用.除此之外,物理信息约束是通过自动微分技术计算波动方程的参数项,其时间复杂与神经网络参数量相关,当神经网络的参数量较大时,模型训练时间会较长.

本文提出一种基于波动方程的有限差分残差约束(Finite Difference Residual Constraints Loss, FDRC Loss)用于训练神经网络.不同于 PINNs 类方法,FDRC Loss 是一种显式约束,它将有限差分方法在一个计算时间步内输入场和输出场的映射关系作为神经网络拟合目标.这种映射关系与偏微分方程的边界条件无关,这使得神经网络能够求解不同边界条件的波动方程,具备了更好的应用潜力.除此之外,FDRC Loss 的计算不依赖自动微分技术,因此其计算时间复杂度与神经网络参数量无关,这降低了训练成本.

我们将神经网络的输出 $\hat{p}$、$\hat{u}$、$\hat{v}$ 视作一个有限差分方法在下一个时间步的解,并基于 FDM Kernel 计算其关于时间和空间的偏导项,将这些偏导项带入波动方程以构建方程残差项,其中 FDRC Loss 是方程组所有有残差项的 L2 范数之和.因此,我们分别构建速度损失项 $L_u$、$L_v$ 和压力损失项 $L_p$ 如公式(4)所示.损失项值的大小描述了神经网络对于方程组的求解精确度,当损失项足够小时,我们认为 $\hat{p}$、$\hat{u}$、$\hat{v}$ 预测得足够精确.值得注意的是,在基于交错网格的有限差分方法中,$p^{t+dt}$ 是由 $u^{t+dt}$ 和 $v^{t+dt}$ 推导获得,因此压力损失项 $L_p$ 是将神经网络预测的 $\hat{p}$、$\hat{u}$、$\hat{v}$ 代入公式(2)中压力控制方程得到.

$$\begin{cases} L_u = \left\| \dfrac{\partial \hat{u}}{\partial t} + \sigma_{x,y} \hat{u} + \dfrac{\partial p}{\rho_0 \partial x} \right\|_2 \\ L_v = \left\| \dfrac{\partial \hat{v}}{\partial t} + \sigma_{x,y} \hat{v} + \dfrac{\partial p}{\rho_0 \partial y} \right\|_2 \\ L_p = \left\| \dfrac{\partial \hat{p}}{\partial t} + \sigma_{x,y} \hat{p} + \rho_0 c^2 \left( \dfrac{\partial \hat{u}}{\partial x} + \dfrac{\partial \hat{v}}{\partial y} \right) \right\|_2 \end{cases} \quad (4)$$

不同于 Spline-PINN[22],FDRC Loss 没有构建单独的边界约束,我们将边界条件和波动方程视作一个整体,把完美匹配层系数 $\sigma$ 视作神经网络的一个输入特征,这使得神经网络对于边界区域和非边界区域的预测具有一致性,避免了物理信息方法存在不同性质的损失项导致模型收敛程度不高的问题.

进一步地,对于一个 $N_x * N_y$ 的交错笛卡尔网格离散化的算例,采用 3.2 章节提出的有限差分卷积核计算波动方程偏导数时,在索引为(i,j)的网格点上,其离散损失项可以表示为公式(5)所示:

$$\begin{cases} L_{u(i,j)} = \dfrac{\hat{u}_{(i,j)}^{t+1} - u_{(i,j)}^t}{\Delta t} + \sigma_{(i,j)} \hat{u}_{(i,j)}^{t+1} + \dfrac{p_{(i,j)}^t - p_{(i-1,j)}^t}{\rho_0 \Delta x} \\ L_{v(i,j)} = \dfrac{\hat{v}_{(i,j)}^{t+1} - v_{(i,j)}^t}{\Delta t} + \sigma_{(i,j)} \hat{v}_{(i,j)}^{t+1} + \dfrac{p_{(i,j)}^t - p_{(i,j-1)}^t}{\rho_0 \Delta y} \\ L_{p(i,j)} = \dfrac{\hat{p}_{(i,j)}^{t+1} - p_{(i,j)}^t}{\Delta t} + \sigma_{(i,j)} \hat{p}_{(i,j)}^{t+1} + \\ \rho_0 c^2 \left( \dfrac{\hat{u}_{(i+1,j)}^{t+1} - \hat{u}_{(i,j)}^{t+1}}{\Delta x} + \dfrac{\hat{v}_{(i,j+1)}^{t+1} - \hat{v}_{(i,j)}^{t+1}}{\Delta y} \right) \end{cases} \quad (5)$$

将所有的损失项合并得到最终的有限差分残差约束如公式(6):

$$L_{total}^{wave} = L_u + L_v + L_p \quad (6)$$

### 3.5 无监督训练策略

本节将介绍如何基于训练池策略进行无监督训练.研究人员发现,基于物理的神经网络模型在训练过程中的迭代数据能够用于新一轮的训练,这在 Spline-PINN 的工作中已有验证[22].在我们的方法中,神经网络的输出是不断变化的波场,这种波场的物理保真度会随着模型的收敛而提高,同时波动方程具有传播性,这使得低真度的波场会随着训练传播出计算域,因此训练池的数据真实性是能够随着模型训练提高,这为我们的无监督训练提供了可行性.与 Spline-PINN 方法不同,我们的训练池缓存的物理性质是不同的,这些物理量是压力场和质点振速度,而不是样条系数这种隐变量,因此我们的数据更加接近于数据驱动的场景.

训练框架分为三个模块:训练池模块、神经网络模块、物理信息模块,它们的交互逻辑如图 2 所示.初始状态的训练

池会随机生成边界条件，并基于该边界条件构建计算域，此时计算域为空，其存储形式为全为零的张量。在一次训练过程中，我们从训练池中随机采样若干个计算域，将计算域的物理场集合$\{u^t, v^t, p^t, \sigma\}$叠加作为神经网络模型的输入，其中模型输出单个时间步的物理变量$\{\Delta u, \Delta v, \Delta p\}$，输入物理场和场变量相加得到$t + dt$时刻的输出物理场$\{u^{t+dt}, v^{t+dt}, p^{t+dt}\}$，最后我们计算 FDRC Loss 优化神经网络模型参数，并将模型输出物理场$\{u^{t+dt}, v^{t+dt}, p^{t+dt}\}$更新到训练池中。对于训练池中的计算域，我们会在每次更新训练池时以小概率进行重置，这使得神经网络模型能够学习到波的完整扩散过程。重复上述训练过程，神经网络模型的预测精度和训练数据的保真度会不断提高。最后我们引入了内存缓存来提高数据采样和数据回写的效率。

基于上述思路我们实现了一个不依赖真解(Ground Truth)的无监督训练策略，这解决了数据驱动方法在数据集获取成本高的问题，并提高了模型的训练效率。

## 3 实验结果与分析

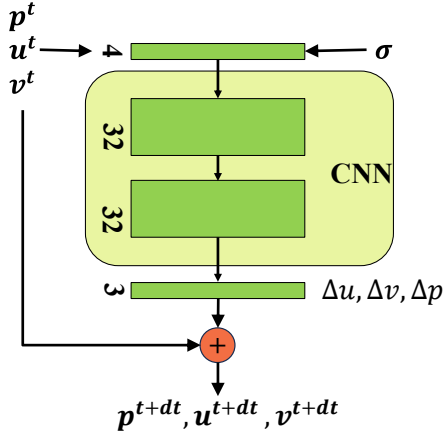

图 4 卷积神经网络结构

Fig. 4 Convolutional neural network structure

### 3.1 实验环境

训练池参数方面，设置计算域数量为 1000，一次采样所选取的样本数(batch size)为 50，取采样 10000 次为一个周期(epoch)。求解方程为无量纲波动方程，其中波速$c=1$，均匀介质密度$\rho_0=1$。所有计算域均被笛卡尔网格划分为 $200 * 200$ 个计算单元，其中计算单元高宽$dx=2$，完美匹配层的厚度为 30 个计算单元宽度，计算域的时间步长$dt=1$。方程的源项$p_{source}$的数量和空间分布随机初始化，并基于于余弦函数构建得到的形式如公式(7)所示，其中$t$是计算域时间，周期$T$和偏执bias随机初始化。

$$p_{source} = \sin\left(\frac{2\pi t}{T} + bias\right) \quad (7)$$

模型参数方面，如图 4 所示卷积神经网络模型由 3 组 3*3

的二维卷积核和$relu$激活函数构成，绿色区域代表每层的通道数量，隐藏层通道数设置为 32。

优化器使用了 Adam 优化器，学习率随 epoch 增加而阶梯式降低，其策略如下：

epoch$\in[0,20]$，学习率为 1e-4；

epoch$\in(20,60]$，学习率为 1e-5；

epoch$\in(60,200]$，学习率为 1e-6。

基于上述训练策略，在 RTX4070 和 6 核 CPU 的计算机上进行训练，每个 epoch 大约耗时 10 分钟，模型约 1~2 天收敛。FDRC Loss 中各项损失变化如图 5 所示，在 20epoch 之前由于学习率较大，各项损失项有较大浮动，当学习率减小后逐渐降低损失项收敛到$10^{-5}$量级。

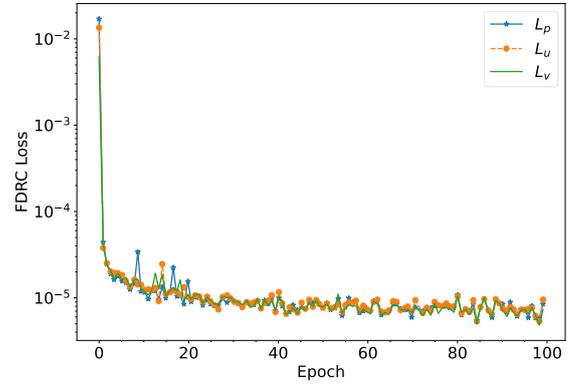

图 5 训练中各个损失项的变化

Fig. 5 The losses at different epoches

### 3.2 模型预测

我们分别模拟了点源传播和多点源传播的物理场景以评价模型对于波动方程的求解能力。对于一个周期$T = 50$的中心源项，其迭代 300 步的计算结果如图 6 所示。图中黑色框线内部为波场传播区域，而黑色框线到计算域边缘的区域为完美匹配层区域，在计算域中心的矩形区域为一个持续震荡的波场源项。观察图 6 左侧神经网络预测结果中，压力场$p$受到中心源项的扰动产生了高压区和低压区，质点振速度$u,v$则产生了高速和低速区，它们均围绕着源源项均匀分布，这说明本文的方法能够学习波动方程内在的波传播物理规律；同时在完美匹配区域中随着不断靠近计算域边缘，$u,v$三个物理场不断衰减，没有产生明显的反射波，这说明神经网络能够学习到完美匹配层的吸波特性。

在波动方程求解中，我们主要关注于源项对计算域压力场的影响，因此重点比较了神经网络(NN)和有限差分方法(FDM)对于压力场$p$的预测。针对单压力源(Single Source)和多压力源(Multiple Source)的传播场景，我们观测了计算时间步$Step \in \{60,120,180,300,50\}$的压力场变化图。图 7 所示。在单压力源的传播场景下，随着时间步推进压力波在计算域中扩散，直到抵达完美匹配层区域(黑框外) 压力波逐渐衰减至

0.与有限差分方法的计算结果对比发现,相同的时间步下神经网络在传播范围、相位、振幅都较为接近有限差分方法;可以看出当压力波传播到完美匹配层区域前,神经网络的预测结果与有限差分方法的十分接近,但随着波传播到完美匹配层区域后,预测波场在第 500 步出现了小幅度扰动,这是因为神经网络在完美匹配层区域的预测存在一定误差,导致了轻微的虚假反射.在多压力源的传播场景下,多个振源相互干扰形成了更加复杂的压力波场,相较于单压力源例,计算域中最大振幅增加,这说明神经网络预测出了波的干涉现象,这也说明了当源项空间位置或数量发生变化时,神经网络能够有效预测,这解决了物理信息神经网络(PINNs)不能改变计算边界条件的问题.

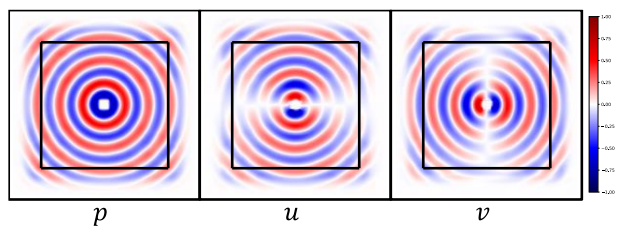

图 6 模型预测的压力场$p$和质点振速场$u,v$

Fig 6. The model predicted pressure field ($p$) and particle velocity fields ($u, v$)

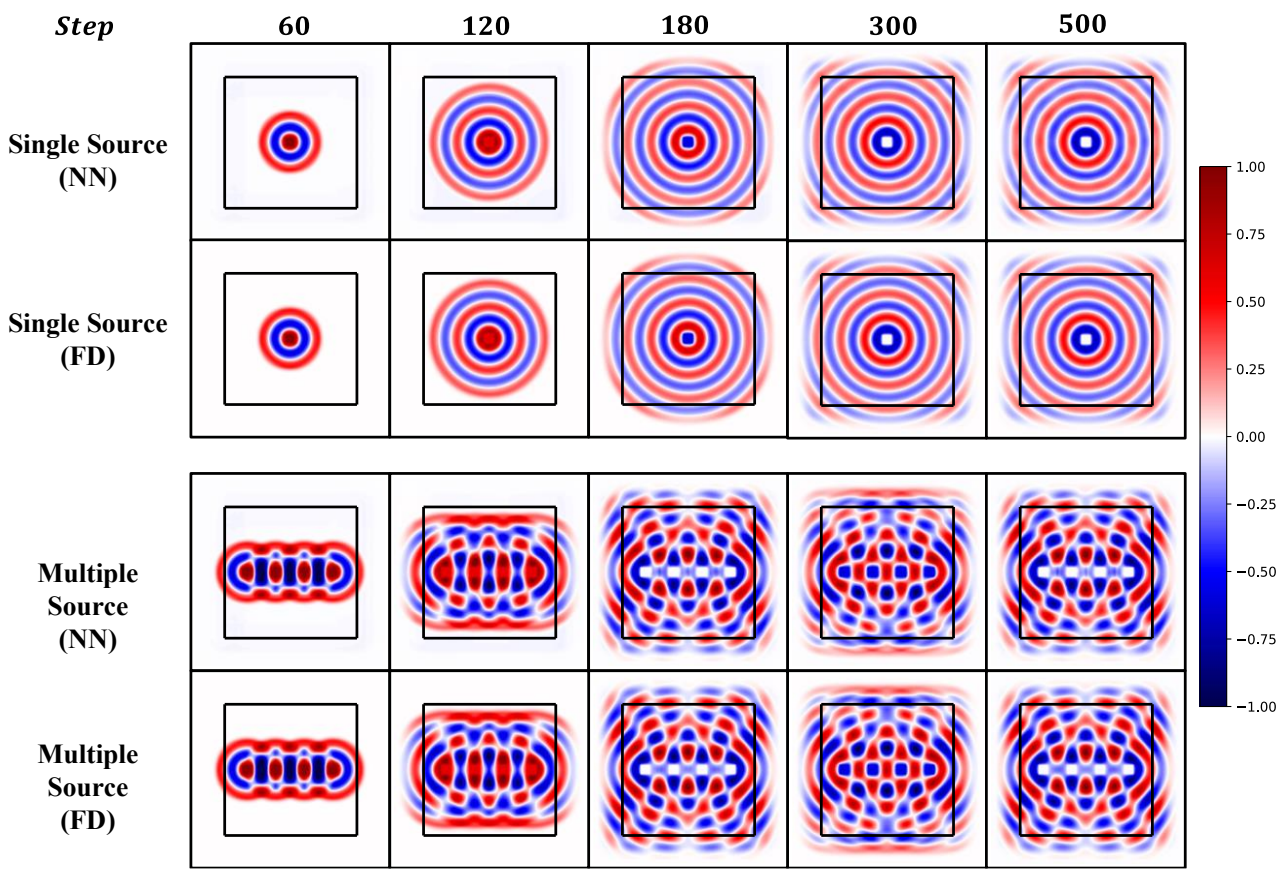

图 7 神经网络模型和有限差分方法对于单压力源和多压力源的波传播预测对比

Fig 7. Physics-based neural networks and finite difference methods for predicting wave propagation from single pressure sources and multiple pressure sources: A comparative analysis

### 3.3 源项泛化

3.2 小节中结果证明了本文方法对于源项空间位置和数量的泛化能力,本小节将继续基于 4.1 节的实验环境评估模型对源项周期变化的泛化能力,并将有限差分的结果作为真解(Ground Truth),以衡量神经网络的预测准确度.通过测试周期$T \in \{10,20,40,60,100\}$的源项在均匀介质中的传播,分别使用有限差分方法(FDM)和神经网络模型(NN)进行模拟,比较二者预测结果的差异.

如图 8 所示,第一行和第二行分别是有限差分方法和神经网络模型对于不同周期源项的模拟结果,第三行是二者的直接差值.可以看出当源项周期$T \in \{20,40,60,100\}$时神经网络模型预测结果十分接近有限差分方法,这说明模型对源项周期变化具有泛化能力,并在合适的离散尺度下保证求解的正确性.然而当源项周期$T = 10$时,预测结果出现了较大误差,这可能是受到了神经网络模型恒定时间步长的限制,当源项周期$T$较小时,恒定的时间步长难以捕捉更高频率的源项变

化,因此在源项周期$T = 10$的算例中,预测压力场在计算域左侧逐渐发散.

上述实验结果证明基于有限差分残差约束的神经网络对于源项周期变化具有泛化能力,这解决了物理信息神经网络(PINNs)关于这方面的问题,这是因为有限差分残差约束的拟合目标是有限差分方法在一个时间步内输入场和输出场的映射关系,而不是去拟合方程的解析解,因此模型能够适应不同源项的波动方程.

图 8 单源项算例中模型对于源项周期变化的预测

Fig 8. The model predicted the periodic variation of the source term in the case of propagation with a single source

### 3.4 误差分析

表 1 神经网络模型在不同计算域中的预测表现

Tab.1 Predictive performance of physical neural networks in different computing domains

| 源项数量 | 源项周期$T$ | 平均损失 | 相对误差$p$ |
|---|---|---|---|
| 1 | 20 | 1.65E-06 | 10.99% |
| 1 | 40 | 1.40E-06 | 9.94% |
| 1 | 60 | 1.47E-06 | 10.55% |
| 1 | 80 | 1.60E-06 | 11.06% |
| 1 | 100 | 1.83E-06 | 11.61% |
| 4 | 20 | 6.70E-06 | 8.27% |
| 4 | 30 | 1.29E-05 | 12.18% |
| 4 | 40 | 4.77E-06 | 7.47% |
| 4 | 50 | 5.16E-06 | 7.87% |

虽然有限差分残差约束的损失值一定程度上反应了模型的收敛情况和预测精度,但这种损失值会随着波场振幅增加而增加,为了更好量化模型的预测精度,将有限差分方法的计算结果作为真值(Ground Truth),并计算神经网络模型预测的相对误差,相对误差如公式(5)所示,其中$X$是预测值,$Y$是真值,$i$是网格采样点的索引,$n$是网格点的总数

量.

$$\text{Mean Relateive Error(MRE)} = \frac{\sum_{i=1}^{n} |X_i - Y_i|}{\sum_{i=1}^{n} |Y_i|} * 100\% \quad (5)$$

基于 4.1 节的实验环境统计训练好的神经网络模型在不同的数值算例中的预测表现如表 1 所示,其中包含了有限差分残差约束(FDRC Loss)的平均损失值和压力场$p$的平均相对误差值(MRE).对于不同的数值算例,FDRC Loss 均处于 1E-06 数量级,其中多压力源的算例中,因为存在干涉的物理现象,波场的平均振幅高于单压力源,因此 FDRC Loss 较高.在相对误差的评估中,不同算例的相对误差值(MRE)表现接近,均在 7%到 12%之间,这说明本文方法对于源项周期和数量的变化具有良好的泛化能力.

### 3.5 方法对比

基于 4.1 节的实验环境,与 Spline-PINN 方法进行了比较,采用相同配置的训练池和相同结构的卷积神经网络进行训练,以避免神经网络模型结构和无监督训练策略带来的影响,最终得到了表 2 的实验结果.表二中分别对比了 Spline-PINN 在一阶和二阶下的表现.通过比较约束的拟合程度差异以及模型训练的效率表现,发现有限差分残差约束在压力损失项和速度损失项均小于 Spline-PINN 方法多个数量级,这证明了有限差分残差约束更容易收敛,这

是因为有限差分残差约束的拟合目的更简单，相对于 Spline-PINN 去拟合波动方程的解析解，学习有限差分方法中单个计算时间步内输入场和输出场的映射关系会更加容易．

在训练效率上，本文的方法在每一轮 epoch(采样 10000 次)耗时为 10min，低于 Spline-PINN 方法在 1 阶的

28.24min 和 2 阶的 64.30min，这是因为 Spline-PINN 在计算物理约束时是基于自动微分技术计算方程偏导项，其计算时间复杂度与神经网络参数量相关，这带来了更多的计算开销和显存占用，而残差分残差约束的计算时间复杂度只与模型输入和输出的物理场大小相关，因此本文方法的训练开销更小．

表 2  与现有方法的比较

Tab.2 Comparison with Existing Methods

| 方法 | 阶数 | 压力损失 | 速度损失 | 训练时间/epoch | 显存占用 | 预测时间 |
|------|------|----------|----------|----------------|----------|----------|
| Spline-PINN | l,m=1 | 8.51E-02 | 1.13E-02 | 28.24min | 6.9GB | 5~7s |
| Spline-PINN | l,m=2 | 5.29E-02 | 6.76E-03 | 64.30min | 7.8GB | 5~7s |
| **Present** | - | **7.35E-06** | **6.8203E-06** | **10min** | **2.4GB** | **2~3s** |

## 4  结论

本文提出了一种基于有限差分残差约束的波动方程无监督学习方法. 基于结构网格和有限差分方法构建卷积神经网络关于波动方程的有限差分残差约束，通过将该约束优化为 0 来训练神经网络，使得神经网络能够预测波动方程在计算域内的正演过程. 基于该约束还提出了一种基于训练池的无监督训练策略，使卷积神经网络能够无数据条件下训练. 实验结果表明，有限差分残差约束相较于 PINNs 类的物理信息约束具有更容易拟合、计算成本更低、边界条件泛化能力更好的优点，同时训练效率更高．

尽管有限差分残差约束对于计算边界条件有着较好的泛化能力，但是对于高频变化源项，神经网络预测误差较大，这说明有限差分残差约束存在时间步长敏感的问题，这也是下一步的研究方向. 此外，我们将进一步将该方法推广应用到三维空间中波动方程中，并进一步探究本文方法的应用潜力．